# Learning Convex Inference of Marginals

**Justin Domke**
Department of Computer Science
University of Maryland
College Park, MD, 20742
domke@cs.umd.edu

## Abstract

Graphical models trained using maximum likelihood are a common tool for probabilistic inference of marginal distributions. However, this approach suffers difficulties when either the inference process or the model is approximate. In this paper, the inference process is first defined to be the minimization of a convex function, inspired by free energy approximations. Learning is then done directly in terms of the *performance* of the inference process at univariate marginal prediction. The main novelty is that this is a direct minimization of empirical risk, where the risk measures the accuracy of predicted marginals.

## 1  Introduction

This paper concerns the learning and inference of marginal distributions. In inference, some vector $\mathbf{y}$ is observed, and the goal is to approximate univariate marginals $p(x_i|\mathbf{y})$ for some true (unknown) distribution $p$, and some hidden, discrete vector $\mathbf{x}$. A typical approach to this problem is to fit the parameters of a graphical model to approximate $p(\mathbf{x}|\mathbf{y})$ with some distribution $q(\mathbf{x}|\mathbf{y})$ using the maximum (conditional) likelihood criterion. Then, in inference, a marginalization algorithm is used to compute $q(x_i|\mathbf{y})$.

This approach is often well-justified– given a correct model, $q(\mathbf{x}|\mathbf{y})$ will converge to $p(\mathbf{x}|\mathbf{y})$ in the high data limit. If exact marginalization is possible, the estimates $q(x_i|\mathbf{y})$ will also converge to the true marginals. However, in practice there are often two major problems.

1. **Computational Intractability**. In general, exact inference is intractable in graphical models with high treewidth (such as "grids"), forcing the use of approximate algorithms. For undirected models, where maximum likelihood learning algorithms require repeated inference, learning will also be intractable. (Maximum likelihood learning of a directed model is usually tractable, even when inference is not.) Even if exact learning is feasible, it is unclear if the results are the best, under *approximate* inference.

2. **Model Defects**. Usually in practice, the model is not exactly correct. (That is, the set of conditional independencies asserted by the graph are not exactly true, or the parametrization of individual factors is imperfect.) This means the parameters *cannot* converge to the "true parameters", since the true distribution is not representable. It is known in this case that maximum likelihood learning will converge to representable distribution with minimum KL-divergence to the true distribution. This is different from the distribution that gives the best predicted marginals, even assuming exact inference.

This paper seeks to ameliorate both of the above issues. This is done by, first, fixing the inference step to be a (presumed tractable) minimization of a convex function. The learning step then consists of fitting the parameters of that function such that the *performance of the inference process* at marginal prediction is best. Specifically, this paper suggests learning the parameters of some function $F(\mathbf{y}, \{b_r(\mathbf{x}_r)\})$. For any observation $\mathbf{y}$, $F$ must be convex over the set of local "pseudomarginals", or "beliefs" $\{b_r(\mathbf{x}_r)\}$.

Since $F$ is convex over $\{b_r(\mathbf{x}_r)\}$, it is nothing more than an implicit mapping from observed variables to beliefs. Thought of as a mapping, it is natural to think that the parameters of $F$ could be adjusted to align these predictions with training data. It turns out that this is true, and moreover, the agreement between predictions and true values can be quantified in terms of a user-specified loss function. This is the real advan-tage of this approach– in learning, the parameters are directly fit to give good marginal predictions, as specified by the loss function. However, a consequence is that only a *mapping* is learned. The approach is not equivalent to approximating the conditional distribution $p(\mathbf{x}|\mathbf{y})$. Nevertheless, predicted marginals are all that are needed for many problems, and experiments suggest this approach can give good results in this case.

## 2 Inference

This paper is based on minimizing a class of convex functions $F$ motivated by free energy approximations (see Section 4).

$$F = \sum_{f \in \mathcal{F}} \sum_{r \in \mathcal{R}} \sum_{\mathbf{x}_r} w_f(\mathbf{x}_r, \mathbf{y}_r) f(b_r(\mathbf{x}_r)) \qquad (1)$$

$\mathcal{F}$ is a set of convex functions of local beliefs. A typical example would be the identity function and the negative entropy function (i.e. $\mathcal{F} = \{b, b \log b\}$). However, in principle, any function that is convex over the interval $(0, 1)$ may be used.

$\mathcal{R}$ is a set of regions. For example, set of of indices of $\mathbf{x}$ in the regions might be the union of the set of cliques $\mathcal{C}$ and the set of individual indices $\mathcal{I}$[1].

The "weighting" functions $w_f$ determine the behavior of $F$. Fitting the model corresponds to fitting these weighting functions.

Given some observation $\mathbf{y}$, the beliefs are given by minimizing $F$.

$$\{b_r^*(\mathbf{x}_r)\} = \arg \min_{\{b_r(\mathbf{x}_r)\}} F(\mathbf{y}, \{b_r(\mathbf{x}_r)\}) \qquad (2)$$

This function needs to be minimized subject to some constraints. In this paper, the beliefs will be constrained to be "locally consistent". If again the regions are cliques $\mathcal{C}$ and individual indices $\mathcal{I}$, the constraints are:

---
[1] By a slight abuse of notation, $\mathbf{x}_r$ and $\mathbf{y}_r$ denote "the variables of $\mathbf{x}$ in region $r$", and "the variables of $\mathbf{y}$ in region $r$", respectively. So, for example, some region might contain only a single variable of $\mathbf{x}$, but multiple variables of $\mathbf{y}$.

$$\forall c \in \mathcal{C}, i \in c, x_i, \quad \sum_{\mathbf{x}_{c \setminus i}} b_c(\mathbf{x}_c) = b_i(x_i) \qquad (3)$$

$$\forall c \in \mathcal{C}, \quad \sum_{\mathbf{x}_c} b_c(\mathbf{x}_c) = 1 \qquad (4)$$

$$\forall i \in \mathcal{I}, \quad \sum_{x_i} b_i(x_i) = 1 \qquad (5)$$

$$\forall c \in \mathcal{C}, \mathbf{x}_c, \quad b_c(\mathbf{x}_c) \geq \mathbf{0} \qquad (6)$$

$$\forall i \in \mathcal{I}, x_i, \quad b_i(x_i) \geq 0 \qquad (7)$$

Since each function in $\mathcal{F}$ is assumed to be convex, $F$ will also be, provided that the weighting functions $w_f$ are positive. (The weights for the linear function $f(b) = b$ need not be constrained). Since the constraints are also convex, this is a convex optimization problem.

It is typical to solve problems like this through the use of "message passing" algorithms. However, in this paper, a more abstract representation will be convenient. Briefly, it is not hard to see that it is possible to transform the problem into the form

$$\mathbf{b}^* = \arg \min_{\mathbf{b}} \sum_{f \in \mathcal{F}} \mathbf{w}_f(\mathbf{y})^T f(\mathbf{b}) \qquad (8)$$

$$\text{such that} \quad A\mathbf{b} = \mathbf{d}$$
$$\mathbf{b} \geq \mathbf{0}.$$

Here, the symbol $\mathbf{b}$ is reused (in boldface) to denote a vector containing all beliefs $b_r(\mathbf{x}_r)$ for all regions $r$ and configurations $\mathbf{x}_r$. Similarly, $\mathbf{w}_f(\mathbf{y})$ represents the vector of weights $w_f(\mathbf{x}_r, \mathbf{y}_r)$ for all regions and configurations. To avoid confusion, boldface ($\mathbf{b}$ or $\mathbf{w}_f(\mathbf{y})$) will always be used when referring to the formulation in Eq. 8, while non-boldface ($b_r(\mathbf{x}_r)$ or $w_f(\mathbf{x}_r, \mathbf{y}_r)$) when referring the the original formulation in Eq. 1.

It is easy to see that each of the three linear constraints (Eqs. 3, 4, and 5) can be handled by setting one row of $A$ and one entry of $\mathbf{d}$.

## 3 Learning

The weighting functions $w_f$, are assumed to depend on some parameters $\boldsymbol{\theta}$. Thus, learning consists of fitting $\boldsymbol{\theta}$. Now, a loss function needs to be specified to quantify the performance of the predicted marginals. Take a set of samples $\{(\hat{\mathbf{x}}, \hat{\mathbf{y}})\}$ drawn from some true (unknown) distribution $p(\mathbf{x}, \mathbf{y})$. Broadly speaking, we would like that the marginals produced by the minimization of $F$ "tend to match" those of $p$. Two ways to quantify this are the univariate "log-loss" and "quad-loss".

## 3.1 Log-loss

Suppose that we would like to minimize the sum of expected KL-divergences between the true marginals, and those produced by the convex optimization. Thus, we would like to choose $F$ by minimizing the "risk"

$$F^* = \arg\min_F \sum_{\mathbf{y}} p(\mathbf{y}) \sum_i \sum_{x_i} p(x_i|\mathbf{y}) \log \frac{p(x_i|\mathbf{y})}{b_i^*(x_i|\mathbf{y},F)}.$$
(9)

Here, $b_i^*(x_i|\mathbf{y}, F)$ denotes the belief obtained for variable $i$ by minimizing $F$ on evidence $\mathbf{y}$. Of course, $p$ is unknown, and so the true risk cannot be minimized. Instead, one can attempt a Monte-Carlo estimate of the true risk, and minimize the "empirical risk".

$$F^* = \arg\min_F -\sum_{\mathbf{y}} p(\mathbf{y}) \sum_i \sum_{x_i} p(x_i|\mathbf{y}) \log b_i^*(x_i|\mathbf{y},F)$$
(10)

$$= \arg\min_F -\sum_{\mathbf{y}} \sum_i \sum_{x_i} p(x_i, \mathbf{y}) \log b_i^*(x_i|\mathbf{y},F)$$
(11)

$$\approx \arg\min_F \sum_{\{(\hat{\mathbf{x}},\hat{\mathbf{y}})\}} \underbrace{-\sum_i \log b_i^*(\hat{x}_i|\hat{\mathbf{y}},F)}_{L_{\text{log}}}$$
(12)

Statistical learning theory studies the conditions under which the approximation represented by Eq. 12 converges in the infinite data limit to that of Eq. 11. However, this paper will not address this.

Now, in general, the minimum in Eq. 12 is found by optimizing with respect to some parameters of $F$. Consider the derivative of $L_{\text{log}}$ with respect to some parameter $\theta_i$ of $F$.

$$\frac{\partial L_{\text{log}}}{\partial \theta_j} = -\sum_i \frac{\partial b_i^*(\hat{x}_i|\hat{\mathbf{y}},F)/\partial \theta_j}{b_i^*(\hat{x}_i|\hat{\mathbf{y}},F)}$$
(13)

The issue of how to calculate $\partial b_i^*(\hat{x}_i|\hat{\mathbf{y}},F)/\partial \theta_j$ will be addressed below.

Kakade et al. (2002) introduced a learning criteria equivalent to $L_{\text{log}}$, and an optimization method for the case of exact inference.

## 3.2 Quad-loss

An alternative criterion would be to try to minimize the expected squared difference between the true marginals and the estimated marginals.

$$F^* = \arg\min_F \sum_{\mathbf{y}} p(\mathbf{y}) \sum_i \sum_{x_i} (p(x_i|\mathbf{y}) - b_i^*(x_i|\mathbf{y},F))^2$$
(14)

Again, this true risk can be approximated with an empirical risk.

$$F^* = \arg\min_F$$
(15)
$$\sum_{\mathbf{y}} \sum_i \sum_{x_i} \left(-2p(x_i,\mathbf{y})b_i^*(x_i|\mathbf{y},F) + p(\mathbf{y})b_i^*(x_i|\mathbf{y},F)^2\right)$$

$$\approx \arg\min_F$$
(16)
$$\sum_{\{(\hat{\mathbf{x}},\hat{\mathbf{y}})\}} \underbrace{\sum_i \left(-2b_i^*(\hat{x}_i|\hat{\mathbf{y}},F) + \sum_{x_i} b_i^*(x_i|\hat{\mathbf{y}},F)^2\right)}_{L_{\text{quad}}}$$

Notice that the second term in Eq. 16 does not depend on the observed data $\{\hat{\mathbf{x}}\}$.

Again, it is not hard to calculate the derivative of $L_{\text{quad}}$ with respect to some parameter $\theta_j$ of $F$.

$$\frac{\partial L_{\text{quad}}}{\partial \theta_j} = 2\sum_i \Big(-\frac{\partial b_i^*(\hat{x}_i|\hat{\mathbf{y}},F)}{\partial \theta_j}$$
(17)
$$+ \sum_{x_i} b_i^*(x_i|\hat{\mathbf{y}},F)\frac{\partial b_i^*(x_i|\hat{\mathbf{y}},F)}{\partial \theta_j}\Big)$$

## 3.3 Derivatives of Beliefs

The above discussion assumes that it is possible to calculate the derivative of the beliefs with respect to the parameters of $F$. (Recall that the weights $\mathbf{w}_f$ are parameterized by some vector $\boldsymbol{\theta}$.) These derivatives are not obvious, given that the beliefs are determined only implicitly by the minimization of $F$. To calculate them, it is first necessary to establish two claims.

**Claim 1**: Let $F(\mathbf{b}, \boldsymbol{\theta})$ be a continuous function such that for all $\boldsymbol{\theta}$, $F$ that has a unique fixed point in $\mathbf{b}$. Define $\mathbf{b}^*(\boldsymbol{\theta})$ such that $\frac{\partial F(\mathbf{b}^*(\boldsymbol{\theta}),\boldsymbol{\theta})}{\partial \mathbf{b}} = \mathbf{0}$. Then,

$$\frac{\partial \mathbf{b}^*(\boldsymbol{\theta})}{\partial \theta_j} = -\Big(\frac{\partial^2 F(\mathbf{b}^*(\boldsymbol{\theta}),\boldsymbol{\theta})}{\partial \mathbf{b}\partial \mathbf{b}^T}\Big)^{-1} \frac{\partial^2 F(\mathbf{b}^*(\boldsymbol{\theta}),\boldsymbol{\theta})}{\partial \mathbf{b}\partial \theta_j}.$$

This claim is essentially a restatement of the (multivariate) Implicit Function Theorem. Here $\frac{\partial^2 F}{\partial \mathbf{b}\partial \mathbf{b}^T}$ denotes the matrix of second partial derivatives of $F$ with respect to the elements of $\mathbf{b}$. Similarly, $\frac{\partial^2 F}{\partial \mathbf{b}\partial \theta_j}$ denotes the vector of partial derivatives of $\partial F/\partial \theta_j$ with respect to the elements of $\mathbf{b}$. Finally, $\partial \mathbf{b}^*/\partial \theta_j$ denotes the vector of derivatives of the elements of $\mathbf{b}^*$, all with respect to $\theta_j$.

This result is not quite adequate to get the derivatives of beliefs, because it does not consider the constraints.

In the following claim, the argument of $(\mathbf{b}^*(\boldsymbol{\theta}), \boldsymbol{\theta})$ to $F$ is dropped for space.

**Claim 2**: Define $\mathbf{b}^*(\boldsymbol{\theta}) \doteq \arg\min_\mathbf{b} F(\mathbf{b}, \boldsymbol{\theta})$, such that $A\mathbf{b} = \mathbf{d}$ for some convex function $F$. Then,

$$\frac{\partial \mathbf{b}^*(\boldsymbol{\theta})}{\partial \theta_j} = (D^{-1}A^T(AD^{-1}A^T)^{-1}AD^{-1} - D^{-1})\frac{\partial^2 F}{\partial \mathbf{b} \partial \theta_j}$$

where $D = (\frac{\partial^2 F}{\partial \mathbf{b} \partial \mathbf{b}^T})$.

A proof is in the appendix. Essentially, the proof consists of augmenting the set of variables with a vector of Lagrange multipliers to enforce that $A\mathbf{b} = \mathbf{d}$, and then applying Claim 1 to the full set of variables $\{\mathbf{b}, \boldsymbol{\lambda}\}$. (See below for the constraint that $\mathbf{b} \geq 0$.)

For the function of interest in this paper, use the formulation of $F$ in Eq. 8.

$$\left(\frac{\partial^2 F}{\partial \mathbf{b} \partial \mathbf{b}^T}\right) = D = \mathrm{diag}(\sum_f \mathbf{w}_f(\mathbf{y}) \odot f''(\mathbf{b})) \qquad (18)$$

Here, $\odot$ denotes element-wise multiplication, and $f''$ denotes the second derivative of $f$. Notice that by virtue of being diagonal, inverting $D$ is trivial and only consists of inverting each entry.

The last term in Claim 2 is also easy to calculate.

$$\frac{\partial F}{\partial \mathbf{b} \partial \theta_j} = \sum_f \frac{\partial \mathbf{w}_f(\mathbf{y})}{\partial \theta_j} \odot f' \qquad (19)$$

Finally, the partial derivatives $\partial \mathbf{w}_f(\mathbf{y})/\partial \theta_j$ are determined directly by $\partial w_f(\mathbf{x}_c, \mathbf{y}_c)/\partial \theta_j$. The exact form, of course, depends on the way in which the weighting functions $w_f(\mathbf{x}_c, \mathbf{y}_c)$ are parametrized. In the common case where each of the weighting functions is fully and independently parametrized,

$$\frac{\partial w_f(\mathbf{x}_c, \mathbf{y}_c)}{\partial \theta_{(f', \mathbf{x}'_c, \mathbf{y}'_c)}} = \delta(f = f', \mathbf{x}_c = \mathbf{x}'_c, \mathbf{y}_c = \mathbf{y}'_c),$$

and so the derivatives $\partial \mathbf{w}_f(\mathbf{y})/\partial \theta_j$ are sparse, binary vectors.

The above discussion did not consider the constraint $\mathbf{b} \geq 0$. For the functions $\mathcal{F}$ used in this paper, it is easy to show that if $\mathbf{b}^* = \arg\min_\mathbf{b} f(\mathbf{b}, \boldsymbol{\theta})$, then $\mathbf{b}^* > 0$. (This is established by forming the Lagrangian enforcing $A\mathbf{b} = \mathbf{d}$ (Eq. 28) and taking the gradient with respect to $\mathbf{b}$, which must be zero.)

## 4  A Free Energy Justification

Suppose there is some true distribution over variables $\mathbf{x}$ and $\mathbf{y}$, where each variable is independent of all other variables, given some set of neighbors. Then, by the Hammersley Clifford theorem, the joint distribution can be represented by

$$p(\mathbf{x}, \mathbf{y}) = \frac{1}{Z} \exp(\sum_c E(\mathbf{x}_c, \mathbf{y}_c)), \qquad (20)$$

where the sum is over the set of all cliques $c$ in the neighborhood graph. It follows that the conditional distribution can be written as a "Conditional Random Field" (Lafferty et al., 2001), i.e.

$$p(\mathbf{x}|\mathbf{y}) = \frac{p(\mathbf{x}, \mathbf{y})}{p(\mathbf{y})} = \frac{1}{Z(\mathbf{y})} \exp(\sum_c E(\mathbf{x}_c, \mathbf{y}_c)), \qquad (21)$$

where $Z(\mathbf{y}) = \sum_\mathbf{x} \exp(\sum_c E(\mathbf{x}_c, \mathbf{y}_c))$. Notice that if there are any cliques that contain only variables in $\mathbf{y}$, they can be dropped from Eq. 21.

Now, consider inference. Some vector $\mathbf{y}$ is observed, and we are interested in the conditional distribution induced over $\mathbf{x}$. One approach is to minimize the KL-divergence between some distribution $b(\mathbf{x})$ and $p(\mathbf{x}|\mathbf{y})$.

$$b^*(\mathbf{x}) = \arg\min_b \sum_\mathbf{x} b(\mathbf{x}) \log \frac{b(\mathbf{x})}{p(\mathbf{x}|\mathbf{y})} \qquad (22)$$

$$= \arg\min_b \sum_\mathbf{x} b(\mathbf{x}) \log b(\mathbf{x}) \qquad (23)$$

$$- \sum_\mathbf{x} \sum_c b(\mathbf{x}) E(\mathbf{x}_c, \mathbf{y}_c) + \sum_\mathbf{x} b(\mathbf{x}) \log Z(\mathbf{y})$$

$$= \arg\min_b \sum_\mathbf{x} b(\mathbf{x}) \log b(\mathbf{x}) \qquad (24)$$

$$- \sum_c \sum_{\mathbf{x}_c} b(\mathbf{x}_c) E(\mathbf{x}_c, \mathbf{y}_c)$$

The second term in Eq. 24, called the average energy, can be computed exactly. However, the first term, known as the entropy, is generally too expensive to compute exactly. In fact, even to represent an arbitrary distribution $b(\mathbf{x})$ will be impossible when the dimension of $\mathbf{x}$ is large. The typical way out of this difficulty is to represent only local marginal distributions, and then to approximate the entropy with a Bethe or Kikuchi approximation (Yedidia et al. 2005). One such approximation would be that

$$\sum_\mathbf{x} b(\mathbf{x}) \log b(\mathbf{x}) \approx \sum_{c \in \mathcal{C}} n_c \sum_{\mathbf{x}_c} b(\mathbf{x}_c) \log b(\mathbf{x}_c) \\ + \sum_{i \in \mathcal{I}} n_i \sum_{x_i} b(x_i) \log b(x_i), \qquad (25)$$

for appropriate "counting numbers" $n_c$ and $n_i$ chosen for each clique and variable, respectively. Specifically, the Bethe approximation consists of choosing $n_c = 1$, and $n_i$ is 1 minus the number of cliques in which node $i$ is contained, i.e. $n_i = 1 - |\{c : i \in c\}|$. More sophisticated approximations consider different regions and different counting numbers.

For a singly-connected graph, the Bethe approximation is exact. For general graphs, it is not always clear what choice of counting numbers will result in a good approximation to the entropy or (more importantly) accurate estimated marginals.

The constraints given in Section 2 (Eqs. 3-7) are, in general, a convex relaxation of the "marginal polytope". In general, given a set of local marginal distributions $\{b_r(\mathbf{x}_r)\}$, no global distribution $b(\mathbf{x})$ that satisfies them all (Yedidia et al. 2005). In order to guarantee that a global distribution exists, one must impose a number of linear constraints, yielding a polytope (Wainwright and Jordan 2003). However, for graphs with high treewidth, the number of constraints is very large, which motivates only enforcing a subset of the constraints. (The local consistency constraints are exact in the case of a tree-structured graph.) This can be a problem since the KL-divergence justification given above assumes that the minimization is over *valid* distributions $b(\mathbf{x})$. There will usually exist a set of inconsistent marginals that have a lower estimated KL-divergence than any true distribution, resulting in less accurate predictions.

The above discussion motivates learning the parameters from data. Rather than attempting to approximate the true $E(\mathbf{x}_c, \mathbf{y}_c)$ (if any), and the true entropy, one could fit the parameters to give the best *predicted marginals*, in light of the relaxation to the marginal polytope, as well as any model defects. This would suggest fitting $E(\mathbf{x}_c, \mathbf{y}_c)$, as well as $n_c$ and $n_i$, where the mapping from the observation $\mathbf{y}$ to the predicted marginals are given by

$$\{b_r^*\} = \arg\min_{\{b_r\}} \sum_c \sum_{\mathbf{x}_c} -b_c(\mathbf{x}_c) E(\mathbf{x}_c, \mathbf{y}_c) + \quad (26)$$

$$\sum_c n_c \sum_{\mathbf{x}_c} b_c(\mathbf{x}_c) \log b_c(\mathbf{x}_c) + \sum_i n_i \sum_{x_i} b_i(x_i) \log b(x_i).$$

In this paper, this is generalized by, rather than taking a constant counting number ($n_c$ or $n_i$) for each region, taking a weight that is allowed to depend on $\mathbf{y}_c$ and $\mathbf{x}_c$. This is not suggested by the free energy approximation, but the experiments below suggest it can enable a more accurate approximation of the marginals. Second, rather than just taking terms with $b$ or $b \log b$, consider arbitrary convex functions of the beliefs. This results in the form for $F$ introduced above,

$$\{b_r^*\} = \arg\min_{\{b_r\}} \sum_{f \in \mathcal{F}} \sum_{r \in \mathcal{R}} \sum_{\mathbf{x}_r} w_f(\mathbf{x}_r, \mathbf{y}_r) f(b(\mathbf{x}_r)). \quad (27)$$

Table 1: Test errors with 30% noise. (All per pixel)

| Method\Error | Classif. | Regress. | $L_{\log}$ | $L_{\text{quad}}$ |
|---|---|---|---|---|
| Pseudo.+M.F. | 0.0475 | 0.0445 | 0.2705 | -0.9109 |
| Pseudo.+B.P. | 0.0495 | 0.0425 | 0.2109 | -0.9149 |
| Mean Field | 0.0300 | 0.0265 | 0.1265 | -0.9470 |
| Belief Prop. | 0.0373 | 0.0314 | 0.1393 | -0.9371 |
| $L_{\log}$ | 0.0261 | 0.0201 | .0694 | -0.9598 |
| $L_{\text{quad}}$ | 0.0260 | 0.0201 | .0723 | -0.9598 |

Table 2: Test errors with 50% noise. (All per pixel)

| Method\Error | Classif. | Regress. | $L_{\log}$ | $L_{\text{quad}}$ |
|---|---|---|---|---|
| Pseudo.+M.F. | 0.0954 | 0.0920 | 0.6291 | -0.8161 |
| Pseudo.+B.P. | 0.0977 | 0.0866 | 0.5056 | -0.8268 |
| Mean Field | 0.0751 | 0.0668 | 0.3030 | -0.8663 |
| Belief Prop. | 0.0783 | 0.0671 | 0.3039 | -0.8658 |
| $L_{\log}$ | 0.0575 | 0.0417 | 0.1384 | -0.9165 |
| $L_{\text{quad}}$ | 0.0561 | 0.0414 | 0.1428 | -0.9172 |

This is also less general than normal free energy approximations, in the sense that the weights for the $f = b \log b$ term are constrained to be positive, while counting numbers can be negative.

## 5 Experiments

As a basic test of this approach, the algorithm was used to learn to "denoise" binary images of handwritten digits from the MNIST database. Ten images of the digits 1-9 were randomly selected for training and testing datasets. Each of the images was then subjected to various amounts of noise. For example, 10% noise means that each pixel has a 10% chance of being assigned randomly. The noisy images make up the observed vectors $\mathbf{y}$, while the original images make up the hidden vectors $\mathbf{x}$.

The model uses regions consisting of individual nodes of $\mathbf{x}$ and $\mathbf{y}$ at the same location, and pairs of neighboring nodes, also at the same locations. The same weights were used for horizontal and vertical regions. The weights were not constrained to be symmetric.

The PDCO primal-dual interior method[2] was used to optimize $F$. Since the functions $f$ are twice differentiable, the full (diagonal) Hessian can be calculated in closed form.

---
[2] www.stanford.edu/group/SOL/software/pdco.html

Table 3: $w_{b \log b}(\mathbf{x}_c, \mathbf{y}_c)$ and $w_{b \log b}(x_i, y_i)$ for $L_{\text{quad}}$ with 50% noise

| $\mathbf{x}_c \backslash \mathbf{y}_c$ | (0,0) | (0,1) | (1,0) | (1,1) |
|---|---|---|---|---|
| (0,0) | 4.86 | 0.04 | 0.05 | 0.02 |
| (0,1) | 4.22 | 3.86 | 4.54 | 5.00 |
| (1,0) | 4.13 | 4.49 | 2.14 | 5.13 |
| (1,1) | 0.06 | 0.02 | 0.03 | 0.02 |

| $x_i \backslash y_i$ | 0 | 1 |
|---|---|---|
| 0 | 4.47 | 0.02 |
| 1 | 0.03 | 0.03 |

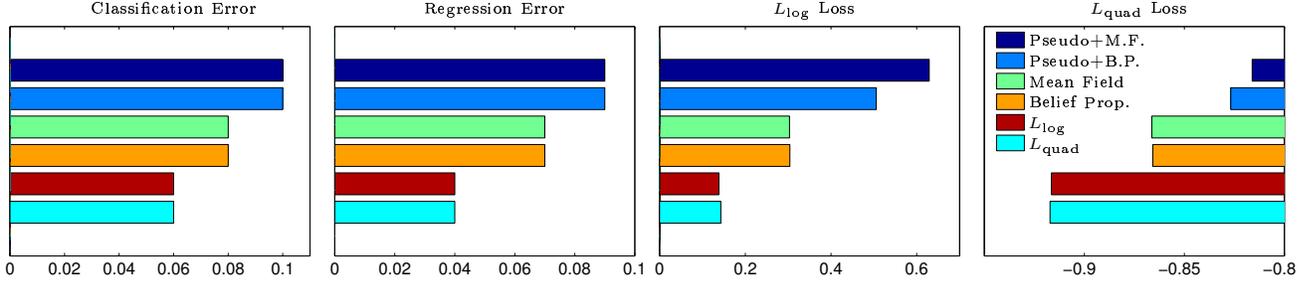

Figure 1: Various measures of accuracy for the six compared methods, with 50% noise.

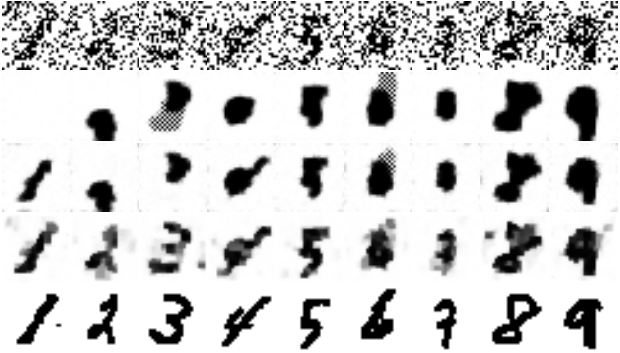

Figure 2: Example results with 50% noise. Top row: Randomly chosen noisy input images of the digits 1-9 from the test set. Second Row: Results from Pseudo.+B.P. Third row: Results from Belief Prop. Fourth Row: Beliefs from convex inference with $L_{\log}$ learning. Bottom row: Original images without noise.

The method prosed here was used, with both loss functions $L_{\log}$, and $L_{\text{quad}}$, using the set of functions $\mathcal{F} = \{b, b \log b\}$. The weighting functions are fully parametrized.

Using the results of Section 3 and the "inner-loop" optimization of $F$ above, it is possible to compute empirical risk $\sum_{\{(\hat{\mathbf{x}}, \hat{\mathbf{y}})\}} L$ and its gradient $\sum_{\{(\hat{\mathbf{x}}, \hat{\mathbf{y}})\}} \partial L / \partial \boldsymbol{\theta}$. An "outer-loop" optimization can use these computation to minimize the empirical risk. Experimentally, small numerical fluctuations in the inner loop can cause instability in the overall optimization. To reduce this, the tolerances on the optimization over $F$ were set very conservatively. Then, a quasi-Newton method (BFGS) was used for the outer optimization.

All weights for the $b$ and $b \log b$ terms are initialized to 0 and 1, respectively. First, 100 iterations are taken, with the weights $w_{b \log b}$ frozen to one, and then another 100 iterations are taken for all weights. This heuristic seems to somewhat improve convergence.

These models are compared against the CRF implementation[3] of Vishwanthan et al. (2006). The features for individual nodes were a constant, and binary variables corresponding to all the possible values of $y_i$ at the same location. In the notation of that paper, $\mathbf{h}_i = [1, \delta(y_i = -1), \delta(y_i = +1)]$. The edge features are similar– a constant, as well as binary features corresponding to each of the four possible configurations of the noisy pixels at the same locations. Four CRF approaches were considered: Pseudolikelihood learning with mean field or belief propagation inference, and mean-field and belief-propagation[4], using those algorithms to approximate marginals for learning and then again for inference. The BFGS method was used for optimization. If the inference method failed to converge after 10,000 iterations, it was stopped at that point.

The approaches are evaluated with four different metrics: the losses $L_{\log}$ and $L_{\text{quad}}$, the "classification error", and the "regression error". Classification error measures the performance in the case that a single prediction (0 or 1) needs to be made for each pixel. (This is done by choosing the value with the highest computed marginal probability.) The classification error is the fraction of predictions that are incorrect. Regression error measures the performance of the model at predicting expected values. The error is the sum of squared differences $(\hat{x}_i - b^*(x_i = 1|\hat{\mathbf{y}}))^2$. Notice that this is different from $L_{\text{quad}}$, as $L_{\text{quad}}$ tries to measure the squared difference of *marginal probabilities*.

Tables 1 and 2 give the test losses for 30% and 50% noise. Figure 1 give a graphical representation of the errors for 50% noise. Here, learning to minimize the losses $L_{\log}$ and $L_{\text{quad}}$ in general perform quite similarly (though each does a slightly better job of minimizing its own loss.) In general, these significantly outperform learning and inference with belief propagation, which in turn outperform pseudolikelihood based learning. Figure 2 gives example outputs where the output be-liefs are visualized with a greyscale intensity.

---
[3] www.cs.ubc.ca/~murphyk/Software/CRF/crf.html
[4] In all cases, "belief propagation" refers to the sum-product formulation.

An interesting question is if allowing the weighting function for the entropy term to vary over the configurations (as opposed to a constant counting number) improves predictions. Table 3 shows the values found for 50% noise. In this particular case at least, the values are highly variant over the configurations, suggesting that this flexibility *is* helpful.

## 6 Related Work

The basic idea that when approximate inference must be used, the learning process should be cognizant of this has been suggested several times. Domingos (2007) suggested "Deep Combination of Learning and Inference" as one of the most important problems problems of the next ten years.

Specific related work demonstrating this principle includes Wainwright (2006) on the value of an inconsistent estimator in the context of approximate inference. This paper creates a tractable surrogate to the entropy. This same surrogate is used for both learning and inference in such a way that the errors of the two processes can cancel each other to some degree – the approximate inference algorithm run on the parameters resulting from the surrogate likelihood can perform better than the algorithm run on true parameters.

Another related area of work is known as "Structured Learning" (Taskar et al., 2004). This is oriented towards MAP inference, rather than marginalization, and is focused on model defects– the objective maximized is given directly in terms of the the maxima of the model, rather than an indirect measure such as the likelihood. Structured learning usually does not focus on computational issues– in a general graph the MAP estimate remains difficult to find. Kulesza and Pereira (2008) provide an analysis of these issues and emphasize the importance of considering the inference process in learning.

## 7 Discussion

Classifiers are often learned by minimizing a loss function closely related to empirical risk. Conversely, graphical models are usually learned by optimizing scores (e.g. the likelihood) rather remote from the performance of the system in the inference stage. This paper presents an approach for learning to infer marginals through a direct minimization of empirical risk, where the risk measures the difference between the true marginals and the marginal predictions of the inference process.

There are several open questions, and promising areas for future research.

In some cases, the marginal beliefs will be used only to predict the maximum probability value for each variable. In such a case, a loss function that penalizes the resulting "labelwise" classification errors (Gross et al., 2007) could be more appropriate than the losses used here.

One advantage of this approach is that essentially the same learning algorithm could be used in the presence of hidden variables. If there are some $x_i$ that are not observed, the sum over the variables in $L_{\log}$ or $L_{\text{quad}}$ can simply be taken over the *observed* variables. This is potentially a significant advantage, since maximum likelihood learning of graphical models requires more advanced methods in the presence of hidden variables.

It is not clear the degree to which constraining the weights for the $f = b \log b$ to be positive term harms the flexibility of the model. One could possibly get more flexibility by only constraining $F$ to be convex over the set of locally consistent marginals, rather than all marginals as done here (Heskes, 2006).

Another possible improvement would be to impose tighter bounds on the marginal polytope than local consistency (Sontag and Jaakkola, 2008), both in the learning and inference step.

Though the generic optimization method used here is reasonably fast, it may be possible to minimize $F$ more efficiently by deriving a message passing algorithm.

Finally, future work could consider generalizing the function $F$. Provided that $F$ is convex and continuous, the results from Section 3.3 will enable learning.

## 8 Acknowledgments

## A Proof

**Claim 2**: Define $\mathbf{b}^*(\boldsymbol{\theta}) \doteq \arg\min_{\mathbf{b}} F(\mathbf{b}, \boldsymbol{\theta})$, such that $A\mathbf{b} = d$ for some convex function $F$. Then,

$$\frac{\partial \mathbf{b}^*(\boldsymbol{\theta})}{\partial \theta_j} = (D^{-1}A^T(AD^{-1}A^T)^{-1}AD^{-1} - D^{-1})\frac{\partial^2 F}{\partial \mathbf{b}\partial \theta_j},$$

where $D = (\frac{\partial^2 F}{\partial \mathbf{b}\partial \mathbf{b}^T})$.

**Proof of Claim 2:** First, create a Lagrangian, enforcing the constraint.

$$\mathcal{L}(\mathbf{b}, \boldsymbol{\theta}) = F(\mathbf{b}, \boldsymbol{\theta}) + \boldsymbol{\lambda}^T(A\mathbf{b} - \mathbf{d}). \qquad (28)$$

Now, apply Claim 1 to the function $\mathcal{L}$ with respect to variables $\mathbf{b}$ and $\boldsymbol{\theta}$. This gives the system

$$\begin{bmatrix} \frac{\partial \mathbf{b}}{\partial \theta_j} \\ \frac{\partial \boldsymbol{\lambda}}{\partial \theta_j} \end{bmatrix} = -\begin{bmatrix} (\frac{\partial^2 \mathcal{L}}{\partial \mathbf{b}\partial \mathbf{b}^T}) & (\frac{\partial^2 \mathcal{L}}{\partial \mathbf{b}\partial \boldsymbol{\lambda}^T})^T \\ (\frac{\partial^2 \mathcal{L}}{\partial \mathbf{b}\partial \boldsymbol{\lambda}^T}) & (\frac{\partial^2 \mathcal{L}}{\partial \boldsymbol{\lambda}\partial \boldsymbol{\lambda}^T}) \end{bmatrix}^{-1} \begin{bmatrix} \frac{\partial^2 \mathcal{L}}{\partial \mathbf{b}\partial \theta_j} \\ \frac{\partial^2 \mathcal{L}}{\partial \boldsymbol{\lambda}\partial \theta_j} \end{bmatrix} \qquad (29)$$

The first submatrix of second partial derivatives is the same as those for the unconstrained system.

$$(\frac{\partial^2 \mathcal{L}}{\partial \mathbf{b}\partial \mathbf{b}^T}) = (\frac{\partial^2 F}{\partial \mathbf{b}\partial \mathbf{b}^T}) \doteq D \qquad (30)$$

The other two matrices are constant.

$$(\frac{\partial^2 \mathcal{L}}{\partial \mathbf{b}\partial \boldsymbol{\lambda}^T}) = A \qquad (\frac{\partial^2 \mathcal{L}}{\partial \boldsymbol{\lambda}\partial \boldsymbol{\lambda}^T}) = 0 \qquad (31)$$

The first derivatives are also easy to calculate.

$$\frac{\partial^2 \mathcal{L}}{\partial \mathbf{b}\partial \theta_j} = \frac{\partial^2 F}{\partial \mathbf{b}\partial \theta_j} \qquad \frac{\partial^2 \mathcal{L}}{\partial \boldsymbol{\lambda}\partial \theta_j} = \mathbf{0} \qquad (32)$$

Substituting all this yields the system

$$\begin{bmatrix} \frac{\partial \mathbf{b}}{\partial \theta_j} \\ \frac{\partial \boldsymbol{\lambda}}{\partial \theta_j} \end{bmatrix} = -\begin{bmatrix} D & A^T \\ A & 0 \end{bmatrix}^{-1} \begin{bmatrix} \frac{\partial^2 F}{\partial \mathbf{b}\partial \theta_j} \\ 0 \end{bmatrix}. \qquad (33)$$

It is possible to recover $\frac{\partial \mathbf{b}}{\partial \theta_j}$ directly from this equation. However, the form derived below can be significantly faster.

Now, consider the inverse of the above matrix. If its entries are $X, Y, Z$ and $U$, by definition it satisfies

$$\begin{bmatrix} D & A^T \\ A & 0 \end{bmatrix} \begin{bmatrix} X & Y \\ Z & U \end{bmatrix} = \begin{bmatrix} I & 0 \\ 0 & I \end{bmatrix}. \qquad (34)$$

Two of the four identities that follow directly from Eq. 34 are

$$\begin{aligned} DX + A^TZ &= I, & (35) \\ AX &= 0. & (36) \end{aligned}$$

Solving Eqs. 35 and 36 for $X$ gives

$$X = D^{-1} - D^{-1}A^T(AD^{-1}A^T)^{-1}AD^{-1}. \qquad (37)$$

The claim follows from the observation that $\frac{\partial \mathbf{b}}{\partial \theta_j} = -X\frac{\partial^2 F}{\partial \mathbf{b}\partial \theta_j}$.$\square$